\documentclass[conference]{IEEEtran}
\IEEEoverridecommandlockouts

\usepackage{cite}
\usepackage{amsmath,amssymb,amsfonts}
\usepackage{algorithmic}
\usepackage{graphicx}
\usepackage{textcomp}
\usepackage{xcolor}
\def\BibTeX{{\rm B\kern-.05em{\sc i\kern-.025em b}\kern-.08em
    T\kern-.1667em\lower.7ex\hbox{E}\kern-.125emX}}

\usepackage{tikz}
\usetikzlibrary{positioning, shapes.multipart, calc, fit, arrows.meta}

\usepackage{subcaption}
\usepackage{eso-pic}

\usepackage{cite}
\usepackage{amsmath,amssymb,amsfonts}
\usepackage{algorithmic}
\usepackage{graphicx}
\usepackage{textcomp}
\usepackage{xcolor}
\usepackage{svg}
\usepackage{multirow}
\usepackage{algorithm}
\usepackage{bm}
\usepackage{booktabs}
\usepackage{todonotes}
\usepackage[font=scriptsize]{caption} 

\usepackage{lipsum}
\usepackage{float}

\begin{document}
\title{Optimized Coordination Strategy for Multi-Aerospace Systems in Pick-and-Place Tasks By Deep Neural Network}

\author{\IEEEauthorblockN{Ye Zhang$^{1,*}$, Linyue Chu$^1$, Letian Xu$^1$, Kangtong Mo$^2$, Zhengjian Kang$^2$ and Xingyu Zhang$^2$}
\IEEEauthorblockA{
$^{1,*}$University of Pittsburgh, PA 15213, USA\\
$^1$University of California, Irvine,  CA 92697, USA\\
$^1$Independent Researcher, USA\\
$^2$University of Illinois Urbana-Champaign, IL 61820, USA\\
$^2$New York University, NY 10012, USA\\
$^2$George Washington University, DC 20052, USA\\
$^\ast$yez12@pitt.edu, $^1$linyuec@uci.edu, $^1$lottehsu@gmail.com\\
$^2$mokangtong@gmail.com, $^2$zk299@nyu.edu, $^2$xingyu\_zhang@gwu.edu}
}

\maketitle

\begin{abstract}
In this paper, we present an advanced strategy for the coordinated control of a multi-agent aerospace system, utilizing Deep Neural Networks (DNNs) within a reinforcement learning framework. Our approach centers on optimizing autonomous task assignment to enhance the system's operational efficiency in object relocation tasks, framed as an aerospace-oriented pick-and-place scenario. By modeling this coordination challenge within a MuJoCo environment, we employ a deep reinforcement learning algorithm to train a DNN-based policy to maximize task completion rates across the multi-agent system. The objective function is explicitly designed to maximize effective object transfer rates, leveraging neural network capabilities to handle complex state and action spaces in high-dimensional aerospace environments. Through extensive simulation, we benchmark the proposed method against a heuristic combinatorial approach rooted in game-theoretic principles, demonstrating a marked performance improvement, with the trained policy achieving up to 16\% higher task efficiency. Experimental validation is conducted on a multi-agent hardware setup to substantiate the efficacy of our approach in a real-world aerospace scenario.
\end{abstract}


\IEEEpeerreviewmaketitle

\section{Introduction}

Space debris has escalated into a critical concern within the global aerospace community, with an estimated 34,000 pieces of debris larger than 10 cm currently orbiting the Earth, posing significant risks to operational satellites and crewed space missions~\cite{gao2023autonomous}. Annually, thousands of fragments are added due to collisions, deteriorating satellites, and remnants of spent rocket stages, further exacerbating the complexity of maintaining safe operational conditions in low Earth orbit (LEO) and beyond. The urgency to address this issue is paramount, as high-velocity collisions can generate extensive debris clouds, significantly multiplying the overall quantity of hazardous fragments. Thus, an effective strategy is required to mitigate the escalating volume of space debris without disturbing the delicate balance of active orbital assets essential for communication~\cite{wang2022orbital}, navigation~\cite{enge2012orbital}, and complex manipulation~\cite{gao2024decentralized}.
\begin{figure}[!t]
    \centering
    \includegraphics[width=0.45\textwidth]{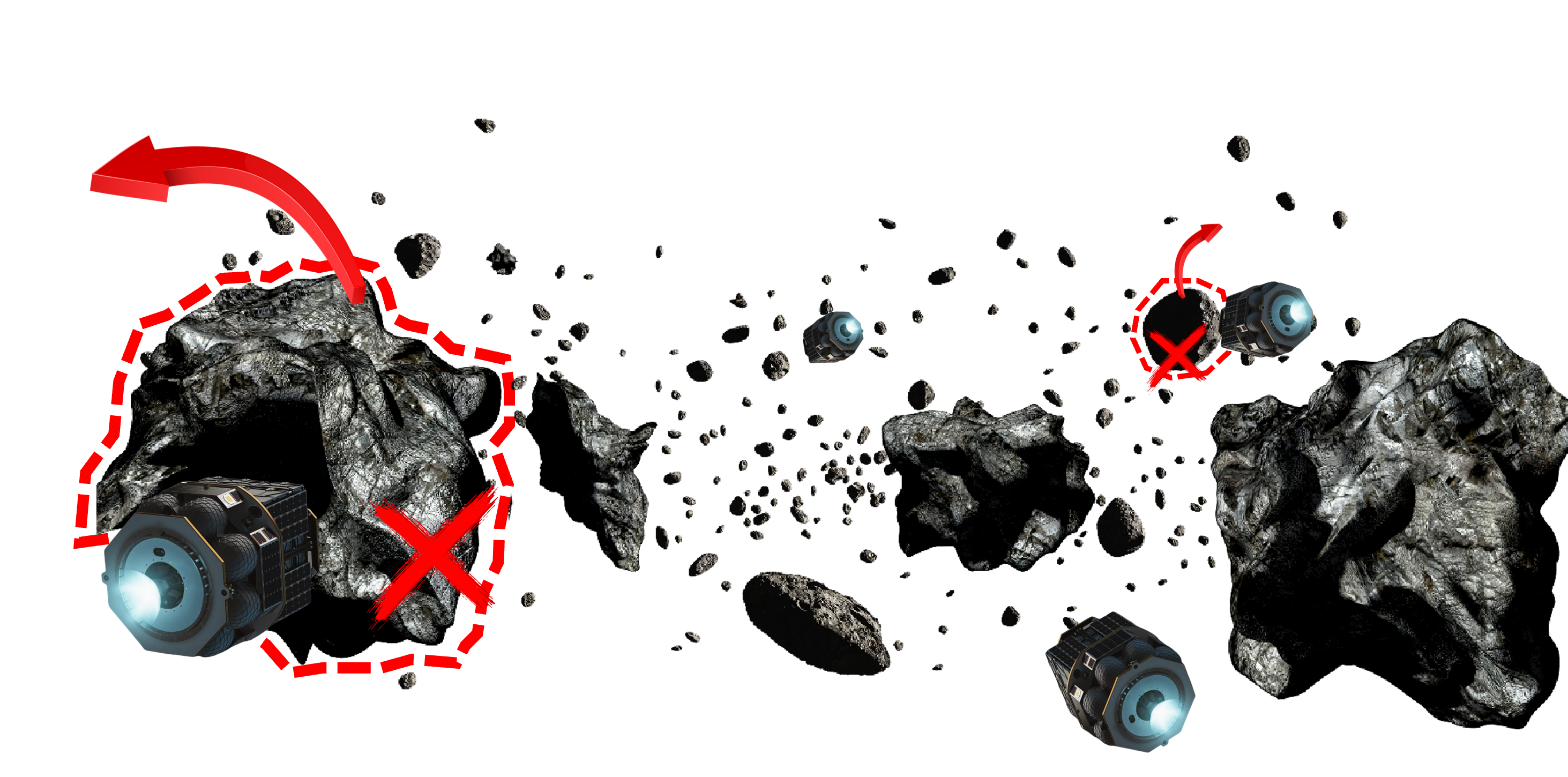}
    \caption{A group of aerospace service modules collaboratively remove the space trash in low Earth orbit(LEO). The red dash-line denotes the aerospace robot has already caught the trash and moved it far away from the LEO.}
    \label{fig:1}
\end{figure}


To achieve efficient debris capture and removal, our research focuses on the development of advanced reinforcement learning-based algorithms that can autonomously assign and optimize pick-and-place (PnP) operations for each robot in the multi-agent system. These algorithms are designed to coordinate debris capture based on each robot’s workspace as shown in Fig.~\ref{fig:1}, relative positioning, and fuel efficiency, ensuring maximized operational longevity and throughput. By framing the debris capture challenge as a reinforcement learning problem, we train a neural network to optimize task assignment under variable conditions, allowing for seamless cooperation among robotic units operating within overlapping or adjacent orbital trajectories.

The complexities of space debris collection necessitate robust, adaptive solutions that can prioritize and allocate tasks based on debris trajectories and density. With reinforcement learning, each robotic unit’s actions are optimized in real-time, enabling dynamic responses to shift debris fields. This paper discusses the implementation of these algorithms~\cite{chen2024tabdeco}, demonstrating their effectiveness through simulation and experimentation in a representative multi-robot hardware setup. The system’s performance is evaluated against a heuristic combinatorial approach, underscoring the potential of our method to significantly enhance task success rates and establish a foundation for scalable, autonomous debris management in space.

\section{Related work}
\subsection{Multi-aerospace robot coordination}
The conventional methodology for evaluating multi-agent aerospace systems often revolves around a singular, global objective function to measure system performance. Historically, substantial research has focused on developing algorithms and strategies within this centralized framework~\cite{keshmiri2009centralized}. Prior studies have explored diverse approaches to multi-agent task allocation, demonstrating efficacy in specific aerospace scenarios. However, these efforts frequently fall short in establishing a truly generalized model for inter-robot collaboration \cite{liao2024inter}. In aerospace task allocation, the challenge is typically framed as the optimal assignment of tasks, such as orbital debris capture or satellite repair, to a fleet of autonomous agents, to minimize an overall cost function while maximizing system robustness and mission longevity.

Traditionally, these frameworks have overlooked the nuanced performance metrics of individual robotic units, concentrating solely on a unified objective function \cite{fini2021unified}. Some studies in dynamic aerospace task allocation have relied on coordination strategies based on local sensory feedback, absent direct inter-agent communication. While these decentralized approaches exhibit scalability and robustness~\cite{wu2024new}, they often lack a holistic perspective that could significantly enhance system-wide performance~\cite{chen2023recontab}. Other strategies employ dynamic programming and heuristic techniques, optimizing task sequences over a finite time horizon to improve mission efficiency. However, such models demonstrate limited adaptability to the unpredictable durations characteristic of real-world aerospace missions, where time horizons are seldom fixed.

Alternatively, common single-agent task prioritization strategies, such as first-in-first-out (FIFO) and shortest-processing-time (SPT), have been adapted for aerospace multi-agent coordination. These rule-based methods are often combined through heuristic search algorithms, such as a greedy search, and are further enhanced for robustness by applying Monte Carlo simulations to manage pattern variability~\cite{mo2024precision}. While this low-complexity, intuitive approach delivers commendable performance and forms a practical baseline, it suffers from notable limitations. Specifically, the lack of inter-agent communication once a task strategy is selected inhibits agents’ situational awareness of one another's operational state. Recognizing the value of enhanced inter-agent information~\cite{wu2024switchtab}, our proposed reinforcement learning-based methodology leverages this latent potential, establishing a more cohesive framework to maximize both individual and collective mission performance.

\subsection{Learning-based multi-aerospace robot coordination}
Reinforcement Learning (RL) has garnered considerable prominence across numerous fields as a potent methodology for enabling autonomous agents to acquire adaptive behaviors within controlled environments. The latest advancements in deep RL have empowered the deployment of complex multi-agent systems, applicable in contexts ranging from competitive multiplayer simulations \cite{zhang2024self} to autonomous vehicular navigation \cite{liu2024enhanced}. Given this progression, there is significant impetus to explore the applicability of these frameworks to multi-aerospace robotic coordination, particularly in the realm of orbital debris management and autonomous satellite maintenance. Recent feasibility studies underscore the substantial promise of RL-driven strategies in facilitating sophisticated cooperation among aerospace agents \cite{zhang2024development}. For instance, Q-learning-based methodologies have been effectively applied to multi-agent scenarios, offering insight into the dynamics of coordinated agent interaction \cite{zhao2024utilizing, zhaoutilizing,yangusing}. These studies reveal the potential of RL to optimize task allocation in spatially distributed aerospace systems by leveraging partially observable environments, where each robotic agent operates with limited knowledge of the global state yet collectively aims to maximize a unified reward function through individual contributions.


The efficacy of this learning-based coordination strategy is benchmarked against established heuristic and rule-based approaches, with results, comparative analyses, and performance metrics presented in Section~\ref{sec::results} and discussed further in Section~\ref{sec::cons}. The findings emphasize the robustness of our RL-based framework in handling the intricacies of multi-agent coordination in the unique context of orbital operations, offering a scalable solution to enhance mission resilience and task completion rates in space environments.

\begin{figure}[!t]
    \centering
    \includegraphics[width=0.35\textwidth]{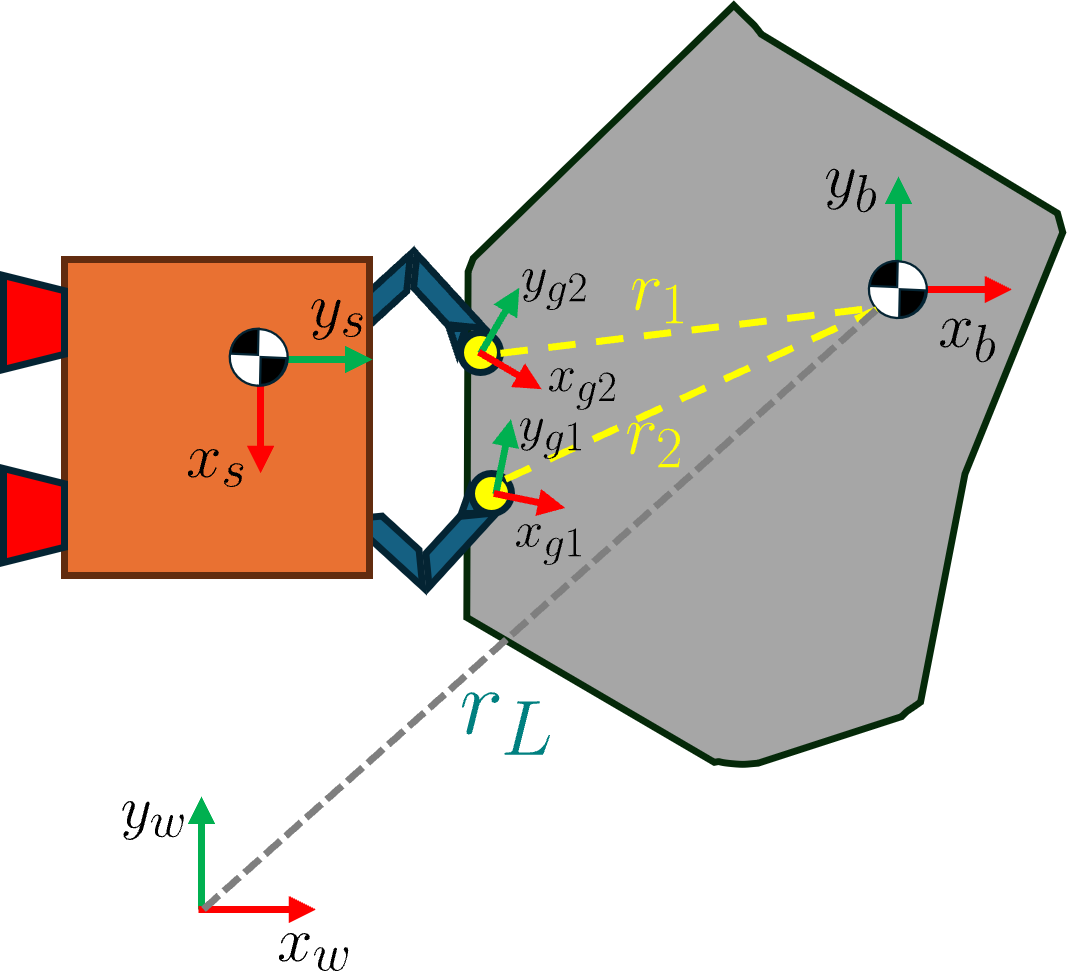}
    \caption{A aerospace service module grasps the debris to transport it in space.}
    \label{fig:2}
\end{figure}

\section{Methodology}
\subsection{Dynamics of the Aerospace Module}
The dual-arm Force/Torque (F/T) decomposition is an essential technique for analyzing and coordinating forces and torques in a dual-arm robotic system engaged in precise load manipulation. The decomposition process enables the system to divide the load's force and torque requirements across both arms, ensuring balanced handling and efficient task execution~\cite{lai2024residual}. The approach relies on defining specific equations for force and moment balance, allowing each arm’s contribution to be independently evaluated relative to the load and its own coordinate frame as shown in Fig.~\ref{fig:2}.

The force exerted by each robotic arm on the load can be represented by the following equation for force balance:

\begin{equation}
    -f_{e1} - f_{e2} + C_L + f_L = m_L \ddot{v}_L
\end{equation}

where $ f_{e1} $ and $ f_{e2} $ are the forces applied by the left and right arms, respectively, $ C_L $ represents the coriolis and centripetal acting on the load,  $ f_L $ is the generalized inertial force acting on the load, $ m_L $ is the mass of the load, $ \ddot{v}_L $ is the acceleration of the load’s center of mass.

\begin{figure*}[!t]
    \centering
    \includegraphics[width=0.9\textwidth]{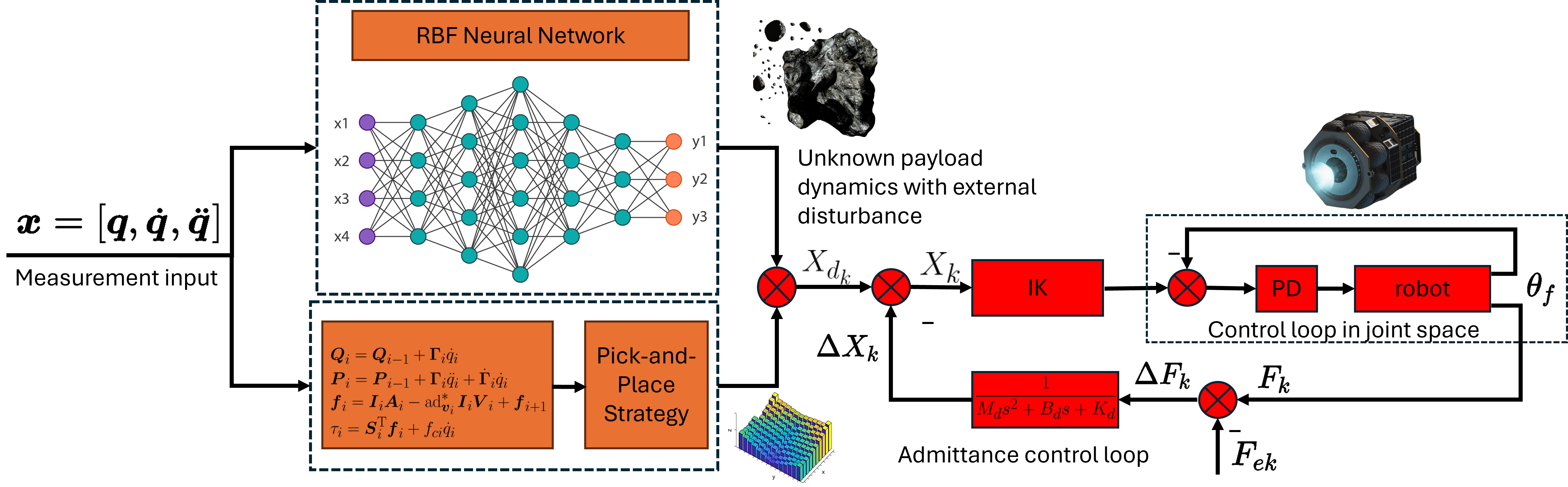}
    \caption{Structure of Neural-Network in pick and place task in for aerospace robot in space.}
    \label{fig:advertising_image}
\end{figure*}

This equation ensures that the sum of forces exerted by both arms, adjusted for gravitational and inertial forces, matches the load's mass and acceleration, achieving a balanced force distribution across both arms.

For moment balance, we consider the torques generated by each arm’s force relative to the center of mass of the load. The moment balance equation can be formulated as:

\begin{equation}
    (\tau_{e1} + r_1 \times f_{e1}) - (\tau_{e2} + r_2 \times f_{e2}) + (\tau_L + r_L \times f_L)
\end{equation}

where $ \tau_{e1} $ and $ \tau_{e2} $ are the torques applied by the left and right arms, respectively. $ r_1 $ and $ r_2 $ are position vectors from the load’s center of mass to the points where each arm exerts its force, and $ r_L $ is the position vector of the load’s center of mass. $ I_L $ represents the moment of inertia of the load. $ \dot{\omega}_L $ is the angular acceleration of the load. And $ \omega_L $ is the angular velocity of the load.

In this equation, the torques from each arm and their respective moments (from the position vectors and force vectors) combine with the load’s own torque. This equation captures both the linear and angular momentum contributions, thus describing the moment balance for the entire system.

By combining these force and moment balance equations, the relationship for the forces and moments applied by both arms in terms of the overall system dynamics can be expressed as:

\begin{equation}
    \begin{bmatrix}
I & 0 \\ 
(r_1)^\times I & I 
\end{bmatrix}
\begin{bmatrix}
-f_{e1} \\ 
-\tau_{e1} 
\end{bmatrix}
+
\begin{bmatrix}
I & 0 \\ 
(r_2)^\times I & I
\end{bmatrix}
\begin{bmatrix}
-f_{e2} \\ 
-\tau_{e2} 
\end{bmatrix}
= 
\begin{bmatrix}
-C_L \\ 
0 
\end{bmatrix}
\end{equation}
where $ I $ is the identity matrix, $ (r_1)^\times $, $ (r_2)^\times $, and $ (r_L)^\times $ denote the cross-product matrices associated with the position vectors $ r_1 $, $ r_2 $, and $ r_L $.

This matrix formulation encapsulates both the force and moment relationships, providing a comprehensive model for the load’s dynamics under the influence of forces and torques from both arms.

To achieve effective control, the robotic system can employ these equations to coordinate each arm’s actions in response to the load’s desired trajectory or motion profile. This approach allows each arm to adjust its force and torque outputs based on real-time feedback, thereby ensuring the load remains balanced and under precise control. The decomposition of the force and torque responsibilities among the arms enables effective load handling and offers flexibility in control design, as each arm can independently adjust to varying load conditions.

\subsection{Newton-Euler Algorithm}
The computation of inverse dynamics for rigid body systems can be efficiently achieved through the Recursive Newton-Euler Algorithm. An adaptation of the RNEA leveraging Lie algebra principles can be formulated to account for the influence of static friction as 

\begin{equation}
\begin{aligned}
& \boldsymbol{Q}_i=\boldsymbol{Q}_{i-1}+\boldsymbol{\Gamma}_i \dot{q}_i \\
& \boldsymbol{P}_i=\boldsymbol{P}_{i-1}+\boldsymbol{\Gamma}_i \ddot{q}_i+\dot{\boldsymbol{\Gamma}}_i \dot{q}_i \\
& \boldsymbol{f}_i=\boldsymbol{I}_i \boldsymbol{A}_i-\operatorname{ad}_{\boldsymbol{v}_i}^* \boldsymbol{I}_i \boldsymbol{V}_i+\boldsymbol{f}_{i+1} \\
& \tau_i=\boldsymbol{S}_i^{\mathrm{T}} \boldsymbol{f}_i+f_{c i} \dot{q}_i
\end{aligned}
\end{equation}

The spatial inertia matrix \( \boldsymbol{I}_i \) and the adjoint transformation matrix \( \text{ad}^*_{\mathbf{V}_i} \) are defined as follows:

\begin{equation}
\boldsymbol{I}_i = 
\begin{bmatrix}
\mathbf{J}_i & m_i \mathbf{p}_i^{\times} \\
- m_i \mathbf{p}_i^{\times} & m_i \mathbf{E}_3
\end{bmatrix}
\end{equation}

\begin{equation}
\mathbf{\Gamma}_i = 
\begin{bmatrix}
\omega_i^{\times} & 0 \\
v_i^{\times} & \omega_i^{\times}
\end{bmatrix}
\end{equation}
where \( \mathbf{V}_i = [\mathbf{v}_i^\top, \, \omega_i^\top]^\top \) and \( \boldsymbol{\Gamma}_i = [\dot{\mathbf{v}}_i^\top, \, \dot{\omega}_i^\top]^\top \) represent the generalized velocity and generalized acceleration, respectively. Here, \( \mathbf{I}_i \) is the spatial inertia. \( m_i \), \( \mathbf{p}_i \), and \( \mathbf{J}_i \) denote the mass, center of mass, and inertia matrix of the \( i \)-th link, respectively. \( (\cdot)^{\times} \) denotes the skew-symmetric matrix, and \( \mathbf{E}_3 \) is the identity matrix.

The term \( f_{ci} \, \text{sgn}(\dot{q}_i) \) is a modification that accounts for the effect of static friction, where \( f_{ci} \) represents the joint static friction coefficient. For a constant revolute joint screw, \( S_i = [0, 0, 0, 0, 0, 1]^\top \).

The differentiable Newton-Euler Algorithm (RNEA) can be represented as a differentiable computational graph, where the inertia parameters and static friction coefficients become learnable parameters. To ensure physically plausible parameters, these dynamic parameters can be reparameterized with virtual parameters that comply with physical constraints~\cite{zhang2020manipulator}.

To guarantee that \( m_i, f_{ci} > 0 \), the mass and static friction coefficients are constructed as follows:

\begin{equation}
\begin{aligned}
   m_i &= \text{exp}(\alpha_i)\\
\quad f_{ci} &= \text{exp}(\beta_i)
\end{aligned}
\end{equation}

where \( \alpha_i \) and \( \beta_i \) are virtual parameters that ensure the positivity of mass and static friction, allowing for stable and realistic dynamic modeling within the framework of the RNEA.

\section{Simulation Results}
\label{sec::results}
The objectives of our investigation are outlined as follows: \textrm{i}) To validate the applicability of our Deep Neural Networks for multi-aerospace robots operating in complex, high-dimensional space environments characterized by intricate dynamics. \textrm{ii}) To demonstrate the advantages of ours method over traditional Nerual Network, e.g., RBFN~\cite{lee2004adaptive}, LSTMs~\cite{yu2019review}, RNNs~\cite{grossberg2013recurrent}, etc, in achieving ultra-high frequency control necessary for precise debris manipulation in orbit. We evaluated our approach using the Isaac Gym platform, a simulation environment specifically tailored for robotics applications, allowing for the emulation of space-based tasks. In this study, the performance of several aerospace robotic systems was examined, with each NN trained over $300$ iterations until convergence, ensuring robust and accurate control for coordinated debris capture and movement within the constraints of an orbital environment.
\begin{figure}[!htp]
    \centering
    \includegraphics[width=0.45\textwidth]{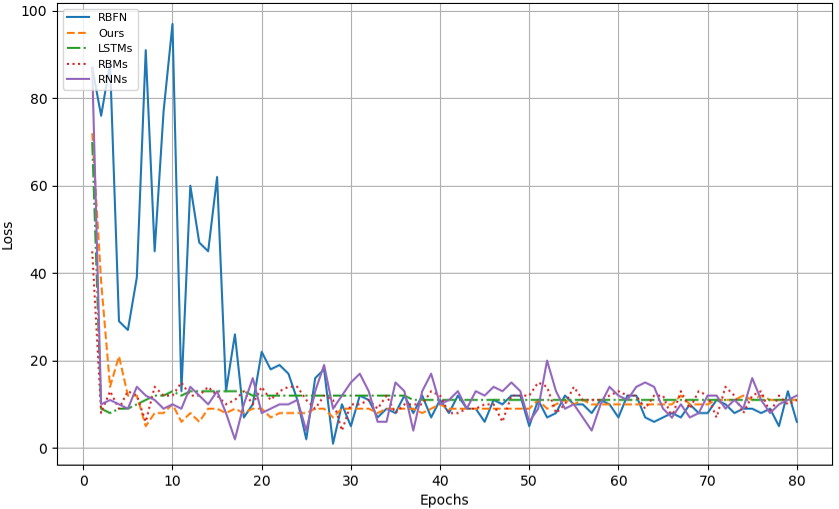}
    \caption{Loss value in task-2 that multi-aerospace robot system picks and places a group of tiny objects in space.}
    \label{fig:advertising_image}
\end{figure}

\begin{figure}[!htp]
    \centering
    \includegraphics[width=0.45\textwidth]{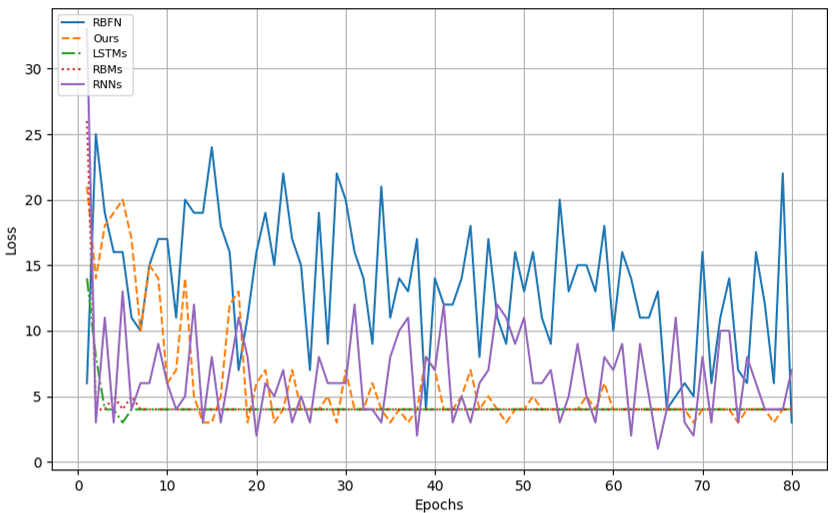}
    \caption{Loss value in task-2 for multi-aerospace robot system pick and place a group of huge object in space.}
    \label{fig:advertising_image}
\end{figure}
We evaluated the proposed methodology on multi-aerospace robots tasked with tracking linear and angular velocity for precise debris manipulation in orbital space. The primary reward was based on achieving accurate velocity tracking (both linear and angular), while penalties were imposed for deviations such as excessive acceleration, unstable positioning, and any incidents that risked mission integrity, such as destabilization of the robotic system. The comparison results for accuracy in each task as shown in Table.~\ref{tab2}

The testing scenarios included various simulated space environments to replicate the challenges of debris collection. For instance, robots were assessed in environments with varying debris distribution patterns, representing dense clusters, dispersed formations, and objects with unpredictable tumbling motions. These scenarios were designed to mimic the dynamic and complex spatial challenges encountered in real orbital debris fields. Each aerospace robot was thus evaluated for its ability to navigate and adjust to these varying spatial configurations, ensuring that the system was capable of handling a diverse range of debris retrieval tasks in space. 

To investigate the advantages of Spiking Neural Networks (SNNs) in high-frequency control scenarios, we increased the simulation time step (\( \Delta t \)) to 2.5 times the standard configuration used in traditional ANN-based tasks, achieving a control frequency of 500 Hz. Conventional ANNs, limited by their energy constraints, typically reach a maximum of 100 Hz, falling short of the motor execution frequency required for optimal performance. In contrast, the energy efficiency of SNNs holds the potential to enhance policy inference speed and facilitate real-time control, making them highly suited for high-frequency tasks. Demonstrating that SNNs can match or surpass ANNs in such high-frequency control applications would establish their superiority for real-world operational environments, especially in contexts requiring rapid response and minimal latency.

\begin{table}[tbp]
\setlength\abovecaptionskip{0.08cm}
\renewcommand{\arraystretch}{1.3}
\caption{Accuracy during the testing process}
\vspace{-0.2cm}
\begin{center}
\begin{tabular}{cccc}
\hline
\specialrule{0em}{0.2pt}{0.1pt}
Method & Epoch($N$=20) & 
Epoch($N$=80) & Epoch($N$=200) \\
\specialrule{0em}{0.1pt}{0.1pt}
\hline
\specialrule{0em}{1pt}{0.1pt}
RBFN & 62.11\% & 72.09\% & 80.32\% \\
LSTMs & 52.65\% & 75.13\% & 86.39\% \\
\specialrule{0em}{-1pt}{-1pt}
RBMs & 78.23.01\% & 85.32\% & 90.25\% \\
RNNs & 73.20\% & 82.98\% & 89.71\% \\
\textbf{Ours} & \boldmath{$82.12\%$} & \boldmath{$92.16\%$} & \boldmath{$95.39\%$} \\
\hline
\end{tabular}
\label{tab2}
\end{center}
\vspace{-0.7cm}
\end{table}

\section{Conclusion}
\label{sec::cons}
We developed an innovative approach for multi-aerospace robot debris retrieval in space environments using a deep neural network. The debris capture task was formulated in simulation, utilizing Poisson disk and grid sampling techniques to model various space debris distributions. Both a novel RL-based strategy and a well-established combinatorial game theory-based approach were designed and implemented within this simulated framework. In tests involving a two-robot configuration, the trained RL policy consistently surpassed the combinatorial method across different debris distributions, achieving up to a 16\% increase in retrieval efficiency. The advantages of the RL method were particularly evident in scenarios with dense clusters of debris, where rapid decision-making and adaptability were critical. Proof-of-concept trials on a two-robot hardware setup successfully validated the approach, demonstrating its potential for real-world application in orbital debris management.

\bibliographystyle{IEEEtran}
\bibliography{reference}

\end{document}